\documentclass{article}
\nonstopmode

\PassOptionsToPackage{numbers, compress}{natbib}

\usepackage[final]{nips_2018}

\usepackage{xspace}
\usepackage[utf8]{inputenc} 
\usepackage[T1]{fontenc}    
\usepackage{hyperref}       
\usepackage{url}            
\usepackage{booktabs}       
\usepackage{amstext}
\usepackage{amsfonts}       
\usepackage{nicefrac}       
\usepackage{microtype}      
\usepackage{comment}
\usepackage{hyperref}

\usepackage{graphicx}

\usepackage{floatrow}
\newfloatcommand{capbtabbox}{table}[][\FBwidth]

\usepackage{subfigure}
\usepackage{authblk}

\usepackage{multirow}

\title{Low Power Inference for On-Device Visual Recognition with A Quantization-Friendly Solution}

\makeatletter
\renewcommand\AB@affilsepx{, \protect\Affilfont}
\makeatother

\author[1]{ Chen Feng}
\author[1]{Tao Sheng}
\author[1]{Zhiyu Liang}
\author[1]{Shaojie Zhuo}
\author[1]{Xiaopeng Zhang}
\author[1]{Liang Shen}
\author[2]{\\Matthew Ardi}
\author[3]{Alexander C. Berg}
\author[4]{Yiran Chen}
\author[5]{Bo Chen}
\author[2]{Kent Gauen}
\author[2]{Yung-Hsiang Lu}

\affil[1]{Qualcomm Inc.}
\affil[2]{Purdue University}
\affil[3]{University of North Carolina at Chapel Hill} 
\affil[4]{Duke University}
\affil[5]{Google Inc.}

\begin{document}

\maketitle

\begin{abstract}
The IEEE Low-Power Image Recognition Challenge (LPIRC) is an annual competition started in $2015$ that encourages joint hardware and software solutions for computer vision systems with low latency and power. Track 1 of the competition in $2018$ focused on the innovation of software solutions with fixed inference engine and hardware. This decision allows participants to submit models online and not worry about building and bringing custom hardware on-site, which attracted a historically large number of submissions. Among the diverse solutions, the winning solution proposed a quantization-friendly framework for MobileNets that achieves an accuracy of $72.67$\% on the holdout dataset with an average latency of $27$ms on a single CPU core of Google Pixel $2$ phone, which is superior to the best real-time MobileNet models 
at the time.
\end{abstract}

\section{Introduction}

Competitions encourage diligent development of advanced  technology. Historical examples include Ansari XPRIZE competitions for suborbital spaceflight, numerous Kaggle competitions such as identifying salt deposits beneath the Earth's surface from seismic images, and the PASCAL VOC, ILSVRC, and COCO competitions for computer vision~\cite{PASCAL,ILSVRC15,COCO}. The IEEE International Low-Power Image Recognition Challenge (LPIRC, \url{https://rebootingcomputing.ieee.org/lpirc}) accelerates the development of computer vision solutions that are low-latency, accurate, and low-power. 

Started in $2015$, LPIRC is an annual competition identifying the best system-level solution for detecting objects in images while using as little energy as possible~\cite{7372672, 8342099, 7858303, arxiv}. Although many competitions are held every year, LPIRC is the only one integrating both computer vision and low power. In LPIRC, a contestants’ system is connected to a referee system through an intranet (wired or wireless). In $2018$, the competition has three tracks. For track $1$, teams submit neural network architectures optimized by Google's TfLite engine and executed on Google's Pixel $2$ phone. For track $2$, teams submit neural network architectures coded in Caffe$2$ and executed on NVIDIA Jetson TX$2$. For track $3$, teams optimize both software and hardware and bring the end system on-site for evaluation. We highlight track 1 below for it has enlisted a large number of high-quality solutions.

\section{Track 1: Efficient Network Architectures for Mobile}

Track 1's goal is to help contestants develop real-time image classification on high-end mobile phones. The platform simplifies the development cycles by providing an automated benchmarking service. Once a model is submitted in Tensorflow format, the service uses TfLite to optimize it for on-device deployment and then dispatches the model to a Pixel 2 device for latency and accuracy measurements. The service also ensures that all models are benchmarked in the same environment for reproducibility and comparability. 

Track 1 selects submissions with the best accuracy within a $30$ms-per-image time constraint, using a batch size of $1$ and a single big core in Pixel 2. Although no power or energy is explicitly measured, latency correlates reasonably with energy consumption. Table \ref{tab:LPIRC$2018$Results} shows the score of the track $1$ winner’s solution. The model is evaluated on both the ILSVRC2012-ImageNet~\cite{ILSVRC15} validation set as well as a freshly collected holdout set.

\begin{table}[htb]
  \caption{Evaluation of Track $1$ Winner’s Solution. \textbf{Average Latency:} is single-threaded, non-batched run-time (ms) measured on a single Pixel $2$ big core of classifying one image. \textbf{Test metric (primary metric):} is the total number of images correctly classified within the wall time ($30$ ms x N) divided by N, where N is the total number of test images. \textbf{Accuracy on Classified:} is the accuracy in $[0, 1]$ computed only with the images classified within the wall-time. \textbf{Accuracy/Time:} is the ratio of the accuracy and either the total inference time or the wall-time, whichever is longer. \textbf{Number Classified:} is the number of images classified within the wall-time.}
  \label{tab:LPIRC$2018$Results}
  \centering
  \begin{tabular}{lccccc}
    \toprule
    \multirow{2}{*} & \shortstack{\textbf{Average}\\\textbf{Latency}} & \shortstack{\textbf{Test}\\\textbf{Metrics}} &\shortstack{\textbf{Accuracy on}\\\textbf{Classified}} & \textbf{Accuracy/Time} & \shortstack{\textbf{Number}\\\textbf{Classified}}\\
    \midrule 
    \textbf{Image Validation} & $28.0$    & $0.64705$  & $0.64705$ & $1.08 \times 10^{-6}$ & $20,000$ \\
    \midrule
    \textbf{Holdout Set} & $27.0$    & $0.72673$  & $0.72673$ & $2.22 \times 10^{-6}$ & $10,927$ \\
    \bottomrule
  \end{tabular}
\end{table}

Track $1$ received a total of $121$ valid submissions (submissions that passed the bazel test and successfully evaluated) and $56$ submissions received test metric scores between $0.59$ and $0.65$. Slightly over half ($51.7$\%) of the solutions use $8$-bit quantization. Most of the  architectures ($74.1$\%) are variations of the existing Mobilenet model family, namely quantized V$2$ ($22.4$\%), quantized V$1$ ($24.1$\%) and float V$2$ ($27.6$\%). The winning track $1$ submission outperformed the previous state-of-the-art below $30$ ms (based on quantized MobileNet V$1$) in accuracy by $3.1$\%. The predominant dependence on Mobilenets is expected considering their exceptional on-device performance and technical support, although future installments are looking to mechanistically discover novel architectures.

\section{Winning Track 1 with Quantization-Friendly Mobilenets}\label{sec:quantizationConvolution}

\subsection{Large Quantization Loss in Precision}

The winning solution is based on MobileNet V$2$, but modified in a way that is quantization-friendly. 
Quantization is often critical for low latency inference on mobile. As most neural networks are trained using floating-point models, they need to be converted to fixed-point in order to efficiently run on mobile devices.
Although Google's MobileNet models successfully reduce the parameter size and computational latency by using separable convolution, direct post-quantization on a pre-trained MobileNet V$2$ model can result in significant precision loss. For example, the accuracy of a quantized MobileNet V$2$ could drop to $1.23$\% on ImageNet validation data-set as shown in Table~\ref{tab:MobileNetResults}. 

\begin{table}[b]
  \centering
  \caption{Experimental results on the accuracy of floating-point and quantization-friendly mobilenets for image recognition and object detection.}
  \label{tab:MobileNetResults}
  \begin{tabular}{ccccc}
    \toprule
     & \multicolumn{4}{c}{\textbf{Image Validation Accuracy}} \\
     & \multicolumn{2}{c}{\textbf{Direct Post-Quantization}} & \multicolumn{2}{c}{\textbf{ Quantization-Friendly}} \\
    \cmidrule(l){2-5}
    \multirow{2}{*}{Model} & \multicolumn{1}{c}{\multirow{2}{*}{Floating-point}} & \multicolumn{1}{c}{\multirow{2}{*}{\shortstack{$8$ Bit\\ Fixed-point}}} & \multicolumn{1}{c}{\multirow{2}{*}{Floating-point}} & \multicolumn{1}{c}{\multirow{2}{*}{\shortstack{$8$ Bit\\ Fixed-point}}}  \\
    & & & & \\
    \cmidrule(l){1-5}
    \multirow{2}{*}{\shortstack{MobileNetV$1$\_$1.0\_224$\\(ImageNet)}} & \multirow{2}{*}{$70.50$\%} & \multirow{2}{*}{$1.80$\%}  & \multirow{2}{*}{$70.77$\%} &  \multirow{2}{*}{$69.88$\%} \\
    & & & & \\
    \midrule
    \multirow{2}{*}{\shortstack{MobileNetV$2$\_$1.0\_224$\\(ImageNet)}}  & \multirow{2}{*}{$71.90$\%}  & \multirow{2}{*}{$1.23$\%} & \multirow{2}{*}{$71.95$\%} & \multirow{2}{*}{$71.01$\%} \\
    & & & & \\
    \midrule
    MobileNetV$1$\_$1.0\_$SSD  & $20.50$\% &  $6.70$\% &  $19.30$\%  &  $17.4$\%  \\
    (COCO)&(COCO $2014$) & (COCO $2014$) & (COCO $2017$) & (COCO $2017$)\\
    \bottomrule
  \end{tabular}
\end{table}

\begin{figure}[htb]
    \centering
    \includegraphics[width=\textwidth]{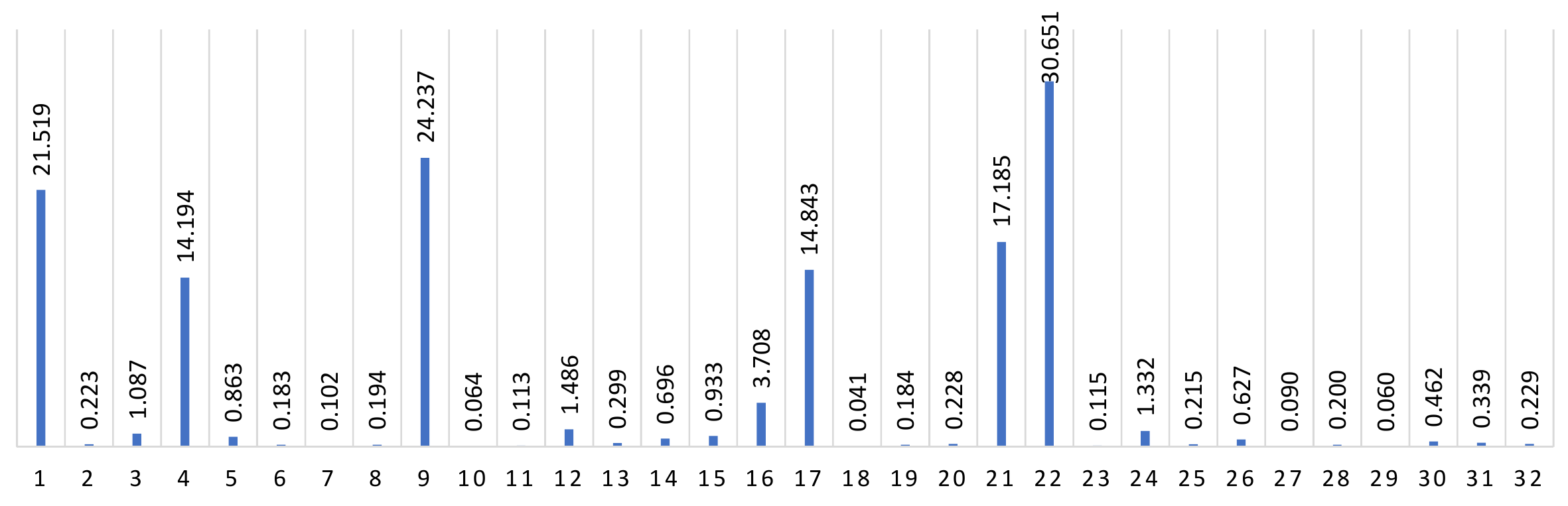}
    \caption{An example of $\alpha$ values across $32$ channels of the first
      depthwise convolution layer from MobileNetV$1$ float model. Small variance in all-zero channels result in large $\alpha$ values. With quantization cross all channels, small $\alpha$ values will suffer huge quantization loss.}
    \label{fig:scaleValues}
\end{figure}

The root cause of accuracy loss due to quantization in such separable convolution networks is analyzed as follows. In separable convolutions, depth-wise convolution is applied on each channel independently, while the min and max values used for weights quantization are taken collectively from all channels. Furthermore, without correlation crossing channels, depth-wise convolution may be prone to produce all-zero values of weights in one channel. All-zero values in one channel have very small variance which leads to a large “scale” value for that specific channel when applying batch normalization transform directly after depth-wise convolution. Therefore, such outliers in one channel may cause a large quantization loss for the whole model due to uneven distributed data range. This is commonly observed in both MobileNet V$1$~\cite{Quantization} and V$2$ models. Figure~\ref{fig:scaleValues} shows an example of the observed batch normalization scale values of $32$ channels extracted from the first depth-wise convolution layer in MobileNetV$1$ float model. As a result, those small values corresponding to informative channels are not well preserved after quantization
and this significantly reduces the representation power of the model.

\subsection{The Winning Quantization-Friendly Approach}
\begin{figure}[htb]
    \centering
    \includegraphics[width=\textwidth]{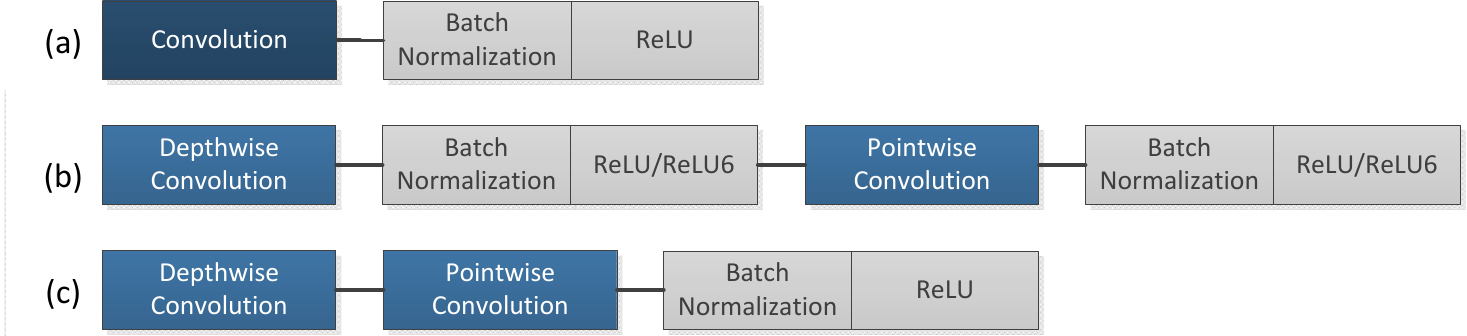}
    \caption{The quantization-friendly separable convolution
      core layer design proposed by the winning solution. (a) and (b) illustrate core layers of the standard convolution and separable convolution. (c) is proposed core layer design based on (b) by removing the batch normalziation and ReLU between depthwise convolution and pointwise convolution.}
    \label{fig:modelArchitecture}
\end{figure}

For a better solution, an effective quantization-friendly separable convolution architecture is proposed as shown in Figure~\ref{fig:modelArchitecture}(c), where the non-linear operations (both batch normalization and ReLU$6$) between depth-wise and point-wise convolution layers are all removed, letting the network learn proper weights to handle the batch normalization transform directly. In addition, ReLU$6$ is replaced with ReLU in all point-wise convolution layers. From the experiments in MobileNet V$1$ and V$2$ models, this architecture maintains high accuracy in the $8$-bit quantized pipeline in various tasks such as image recognition and object detection. 

As an alternative, one can use Learn2Compress~\cite{learn2compress}, Google’s ML framework for directly training efficient on-device models from scratch or an existing TensorFlow model by combining quantization along with other techniques like distillation, pruning, and joint training. Comparing with these options, the winners' solution provides a much simpler way to modify separable convolution layers and make whole network quantization-friendly without re-training.

\subsection{System Integration and Experimental Results}\label{sec:systemIntegration}

By considering the trade-off between accuracy and model complexity, MobileNetV$2$\_$1.0$\_$128$ is chosen as the base architecture to apply the quantization-friendly changes. Based on the proposed  structure, a floating-point model can be trained on the dataset. During the post-quantization step, the model runs against a range of different inputs, one image in each class category from the training data, to collect min and max values as well as the data histogram distribution at each layer output. Values for optimal “step size” and “offset”, represented by $\Delta$ and $\delta$, that minimize the summation of quantization loss and saturation loss during a greedy search, are picked for linear quantization. Given the calculated range of min and max values, TensorFlow Lite provides a path to convert a graph model (.pb) to tflite model (.tflite) that can be deployed on edge devices. 

The proposed fixed-point model with an input resolution of 128 can achieve an accuracy of $64.7$\% on ImageNet validation dataset and an accuracy of $72.7$\% on holdout dataset. The base network model can also be used for different tasks such as image recognition and object detection. Table~\ref{tab:MobileNetResults} shows our experimental results on the accuracy of the proposed quantization-friendly MobileNets  for image recognition on ImageNet (with an input resolution of $224$) and object detection on COCO dataset. With the proposed core network structure, the model largely increases accuracy on both tasks in $8$ bit fixed-point pipeline. Whereas direct quantization on pre-trained MobileNet V$1$ and V$2$ model would cause unacceptable accuracy loss, the quantization-friendly MobileNets managed to stay within $1\%$ of the float model's accuracy for ImageNet classification, and within $2\%$ for COCO object detection.

\section{Conclusions}\label{sec:conclusions}

By providing an convenient platform for evaluating on-device neural network architectures, LPIRC has successfully enlisted creative solutions from the field. Not only does the winning solution outperformed state-of-the-art, it also provides insights regarding quantization that is applicable to other tasks such as object detection. This success showcases not only the effectiveness of their quantization-friendly approach, but also the importance of the platform that facilitates its development.

\bibliography{main}
\bibliographystyle{unsrt}

\end{document}